%% file: anonymous-submission-latex-2023.tex
\documentclass[letterpaper]{article} 
\usepackage{arxiv}  
\usepackage{times}  
\usepackage{helvet}  
\usepackage{courier}  
\usepackage[hyphens]{url}  
\usepackage{graphicx} 
\usepackage{natbib}  
\usepackage{caption} 
\usepackage{algorithm}
\usepackage{algorithmic}

\usepackage[colorlinks,linkcolor=black,citecolor=black]{hyperref}

\usepackage{newfloat}
\usepackage{listings}
\DeclareCaptionStyle{ruled}{labelfont=normalfont,labelsep=colon,strut=off} 
\floatstyle{ruled}
\newfloat{listing}{tb}{lst}{}
\floatname{listing}{Listing}
\usepackage{amsmath,amssymb} 
\usepackage{booktabs}
\usepackage{pifont}
\usepackage{float,multirow,makecell,subfigure,setspace}
\usepackage{threeparttable}
\usepackage{bm}
\usepackage{color}

\newcommand{\ie}{\textit{i.e.}}
\newcommand{\eg}{\textit{e.g.}}
\newcommand{\etal}{\textit{et~al.}}
\newcommand{\xmark}{\ding{55}}%
\newcommand{\rf}[1]{{\textbf{\color{red}{#1}}}} 
\newcommand{\bd}[1]{{\color{blue}{\underline{#1}}}} 

\title{Exploring CLIP for Assessing the Look and Feel of Images}
\author{
    Jianyi Wang\,\,\,\,
	Kelvin C.K. Chan\,\,\,\,
	Chen Change Loy\footnote{Corresponding author.}\\
}
\affiliations{
	S-Lab, Nanyang Technological University \quad \\
	\{jianyi001, chan0899, ccloy\}@ntu.edu.sg
}

\begin{document}

\maketitle

\begin{abstract}
Measuring the perception of visual content is a long-standing problem in computer vision. Many mathematical models have been developed to evaluate the \textit{look} or quality of an image. Despite the effectiveness of such tools in quantifying degradations such as noise and blurriness levels, such quantification is loosely coupled with human language. When it comes to more abstract perception about the \textit{feel} of visual content, existing methods can only rely on supervised models that are explicitly trained with labeled data collected via laborious user study. In this paper, we go beyond the conventional paradigms by exploring the rich visual language prior encapsulated in Contrastive Language-Image Pre-training (CLIP) models for assessing both the quality perception (\textit{look}) and abstract perception (\textit{feel}) of images without explicit task-specific training. In particular, we discuss effective prompt designs and show an antonym prompt pairing strategy to harness the prior. We also provide extensive experiments on controlled datasets and Image Quality Assessment (IQA) benchmarks. Our results show that CLIP captures meaningful priors that generalize well to different perceptual assessments. Our code is publicly available at \url{https://github.com/IceClear/CLIP-IQA}.
\end{abstract}

\input{section/1_introduction}
\input{section/3_methodology}

\input{section/4_discussion}
\input{section/5_conclusion}

\twocolumn
\section*{Acknowledgement}
This research is supported by the National Research Foundation, Singapore under its AI Singapore Programme (AISG Award No: AISG2-PhD-2022-01-033[T]), the RIE2020 Industry Alignment Fund Industry Collaboration Projects (IAF-ICP) Funding Initiative, as well as cash and in-kind contribution from the industry partner(s).
It is also partially supported by the NTU NAP grant.
We thank Kede Ma, Yuming Fang and Hanwei Zhu for providing valuable technical details of SPAQ dataset.

\appendix
\section*{Appendix}

\input{section/2_relatedworks}

\begin{table}[tbp]
    \newcommand{\tabincell}[2]{\begin{tabular}{@{}#1@{}}#2\end{tabular}}
    \renewcommand\arraystretch{1.2}
	\begin{center}
		\caption{Prompt settings for attributes in Sec.~\ref{CLIP_quality} of the main paper}
		\label{table_prompt_32}
		\resizebox{0.48\textwidth}{!}{
		\begin{tabular}{c|c}
            \hline ~\textbf{Attribute}~ & ~\textbf{Prompt pair}~ \\
            \hline \textit{brightness} & [``\texttt{Bright photo.}'', ``\texttt{Dark photo.}''] \\
            \hline \textit{noisiness} & [``\texttt{Clean photo.}'', ``\texttt{Noisy photo.}''] \\
            \hline \textit{colorfulness} & [``\texttt{Colorful photo.}'', ``\texttt{Dull photo.}''] \\
            \hline \textit{sharpness} & [``\texttt{Sharp photo.}'', ``\texttt{Blurry photo.}''] \\
            \hline \textit{contrast} & \makecell[c]{[``\texttt{High contrast photo.}'', ``\texttt{Low contrast photo.}'']} \\
            \bottomrule
		\end{tabular}
		}
	\end{center}
\end{table}

\section{Experimental Settings}
In this section, we provide detailed settings of our experiments in Sec.~\ref{CLIP_quality} and Sec.~\ref{CLIP_abstract} of the main paper. Note that we adopt ResNet-50 based CLIP-IQA for all experiments.

\noindent\textbf{Settings for overall quality.} In Sec.~\ref{CLIP_quality} of the main paper, we compare the performance of CLIP-IQA on assessing the overall quality of images with existing NR-IQA methods.
The comparison is conducted on three widely used NR-IQA datasets, namely KonIQ-10k~
\cite{hosu2020koniq}, LIVE-itW~\cite{ghadiyaram2015massive} and SPAQ~\cite{fang2020perceptual}.
For KonIQ-10k, the training and test set are identical to Hosu~\etal~\cite{hosu2020koniq}.
For LIVE-itW, we follow Hosu~\etal~\cite{hosu2020koniq} to test on the
entire LIVE-itW dataset.
For SPAQ, following Ke~\etal~\cite{ke2021musiq}, we randomly select 20\% of the images for test, and resize the images such that the shorter side has a length of 512.
Note that our test set selection on SPAQ is different from Ke~\etal~\cite{ke2021musiq} since the names of test images are not provided. All the results reported in Sec.~\ref{CLIP_quality} of the main paper are evaluated on our test set for fair comparison.

For non-learning based methods, we directly take the numbers from Hosu~\etal~\cite{hosu2020koniq} for LIVE-itW and KonIQ-10k.
Since no results on SPAQ are available in Hosu~\etal~\cite{hosu2020koniq}, we test these methods on SPAQ using their corresponding official code\footnote{\href{https://github.com/dsoellinger/blind_image_quality_toolbox}{https://github.com/dsoellinger/blind\textunderscore image\textunderscore quality\textunderscore toolbox}}.

\begin{table}[ht!]
    \newcommand{\tabincell}[2]{\begin{tabular}{@{}#1@{}}#2\end{tabular}}
    \renewcommand\arraystretch{1.2}
	\begin{center}
		\caption{Prompt settings for attributes in Sec.~\ref{CLIP_abstract} of the main paper}
		\label{table_prompt_33}
		\resizebox{0.48\textwidth}{!}{
		\begin{tabular}{c|c}
            \hline ~\textbf{Attribute}~ & ~\textbf{Prompt pair}~ \\
            \hline \textit{complex/simple} & [``\texttt{Complex photo.}'', ``\texttt{Simple photo.}''] \\
            \hline \textit{natural/synthetic} & [``\texttt{Natural photo.}'', ``\texttt{Synthetic photo.}''] \\
            \hline \textit{happy/sad} & [``\texttt{Happy photo.}'', ``\texttt{Sad photo.}''] \\
            \hline \textit{scary/peaceful} & [``\texttt{Scary photo.}'', ``\texttt{Peaceful photo.}''] \\
            \hline \textit{new/old} & [``\texttt{New photo.}'', ``\texttt{Old photo.}''] \\
            \bottomrule
		\end{tabular}
		}
	\end{center}
\end{table}

\begin{table*}[ht!]
    \renewcommand\arraystretch{1.1}
	\begin{center}
		\caption{Comparison to NR-IQA methods on synthetic dataset TID2013~\cite{ponomarenko2015image}. Learning-based methods are trained on KonIQ-10k~\cite{hosu2020koniq} and tested on TID2013. \rf{Red} color represents the best performance}
		\label{table_tid}
		\begin{tabular}{c|c|c|c|c|c|c|c|c}
            \toprule \multirow{2}{*}{Metric} &  \multicolumn{4}{c|}{w/o task-specific training} & \multicolumn{4}{c}{w/ task-specific training} \\
            \cline{2-9} & BIQI & BRISQUE & NIQE & \textbf{CLIP-IQA} & KonCept512 & HyperIQA & MUSIQ & \textbf{CLIP-IQA$^+$} \\
            \hline ~SROCC\,$\uparrow$~ & 0.402 & 0.444 & 0.321 & \rf{0.51} & 0.281 & 0.406 & 0.578 & \rf{0.632} \\
            \hline ~PLCC\,$\uparrow$~ & 0.432 & 0.472 & 0.527 & \rf{0.571} & 0.410 & 0.481 & 0.683 & \rf{0.692} \\
            \bottomrule
		\end{tabular}
	\end{center}
\end{table*}

For learning-based methods except for MUSIQ, we follow Hosu~\etal~\cite{hosu2020koniq} to train the models on KonIQ-10k dataset and test on the three datasets.
These methods are trained based on the default settings of the corresponding official code.
Since the training code of MUSIQ is unavailable, we adopt the official model\footnote{\href{https://github.com/google-research/google-research/tree/master/musiq}{https://github.com/google-research/google-research/tree/master/musiq}} trained on KonIQ-10k.

\noindent\textbf{Settings for fine-grained quality.}
We adopt five paired prompts for the fine-grained quality attributes. The prompts in our experiments are shown in Table~\ref{table_prompt_32}.

\noindent\textbf{Settings for abstract quality.}
The prompts used for abstract perception are shown in Table~\ref{table_prompt_33}.

\noindent\textbf{Metrics for overall quality evaluation.}
To evaluate the performance of our CLIP-IQA, we adopt Spearman’s Rank-order Correlation Coefficient (SROCC) and Pearson’s Linear Correlation Coefficient (PLCC), which are widely used for evaluating the performance of IQA methods \cite{bosse2017deep,su2020blindly,zhu2020metaiqa,ke2021musiq,pan2022vcrnet}. Higher SROCC and PLCC indicate more consistency with human-labeled MOS scores.

\noindent\textbf{Implementation.}
We implement our CLIP-IQA with PyTorch and run CLIP-IQA on one NVIDIA Tesla V100 GPU. Code will be made publicly available.

\section{Overall Quality Assessment on Synthetic Data}
In addition to the three real-world IQA datasets in Table 1 of the main paper, we further verify our proposed method on synthetic data.
Here, we choose TID2013 \cite{ponomarenko2015image}, which is one of the most commonly used synthetic IQA datasets.
As shown in Table~\ref{table_tid}, our proposed CLIP-IQA outperforms all other non-learning methods in terms of SROCC and PLCC. With our proposed fine-tuning scheme, CLIP-IQA$^+$ surpasses learning-based methods on TID2013.

\section{Synthetic Samples for Fine-grained Quality}
\begin{figure*}[htb!]
	\begin{center}
		\centerline{\includegraphics[width=2.0\columnwidth]{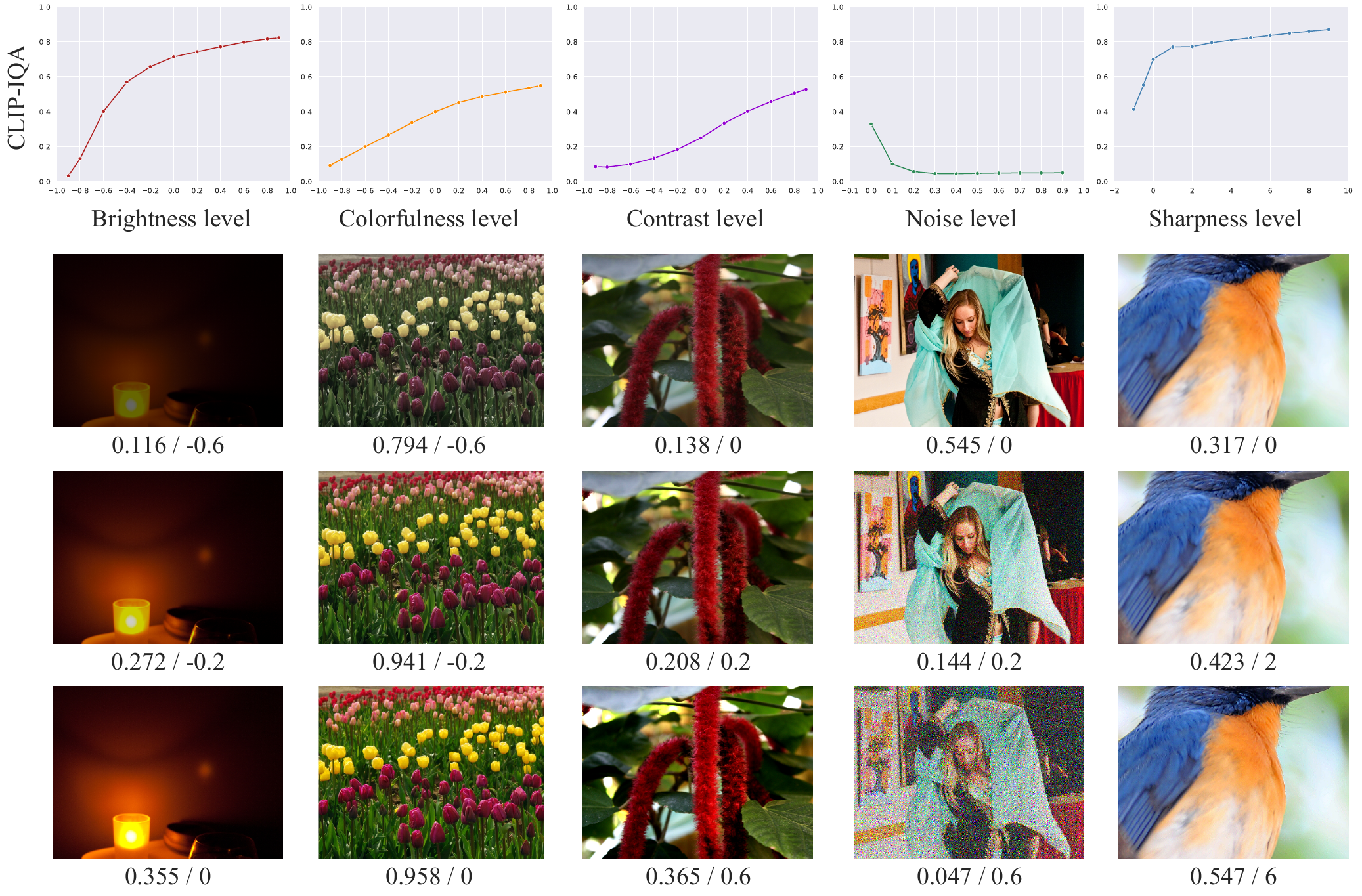}}
		\caption{CLIP-IQA for fine-grained attributes on synthetic data with different input scales. CLIP-IQA clearly demonstrates positive correlations to the change of attributes. The numbers under each image represent the CLIP-IQA score and the extent of each attribute. \textbf{(Zoom-in for best view)}}
		\label{figure_synthetic_supp}
	\end{center}
\end{figure*}

Due to the space limit, no image samples are presented in Fig. 4 of our main paper. Here, we add additional image samples for better illustration. As shown in Fig. \ref{figure_synthetic_supp}, our CLIP-IQA is able to identify fine-grained qualities on synthetic data.

\begin{table}[h]
    \newcommand{\tabincell}[2]{\begin{tabular}{@{}#1@{}}#2\end{tabular}}
    \renewcommand\arraystretch{1.2}
	\begin{center}
		\caption{Prompt settings for additional attributes on AVA~\cite{murray2012ava}}
		\label{table_prompt_ava}
		\resizebox{0.48\textwidth}{!}{
		\begin{tabular}{c|c}
            \hline ~\textbf{Attribute}~ & ~\textbf{Prompt pair}~ \\
            \hline \textit{warm/cold} & [``\texttt{Warm photo.}'', ``\texttt{Cold photo.}''] \\
            \hline \textit{real/abstract} & [``\texttt{Real photo.}'', ``\texttt{Abstract photo.}''] \\
            \hline \textit{beautiful/ugly} & [``\texttt{Beautiful photo.}'', ``\texttt{Ugly photo.}''] \\
            \hline \textit{lonely/sociable} & [``\texttt{Lonely photo.}'', ``\texttt{Sociable photo.}''] \\
            \hline \textit{relaxing/stressful} & \makecell[c]{[``\texttt{Relaxing photo.}'', ``\texttt{Stressful photo.}'']} \\
            \bottomrule
		\end{tabular}
}
	\end{center}
\end{table}

\section{More Quantitative Results for Abstract Perception}
\begin{figure*}[h]
    \setlength{\abovecaptionskip}{-1.5cm}
	\begin{center}
		\centerline{\includegraphics[width=2.0\columnwidth]{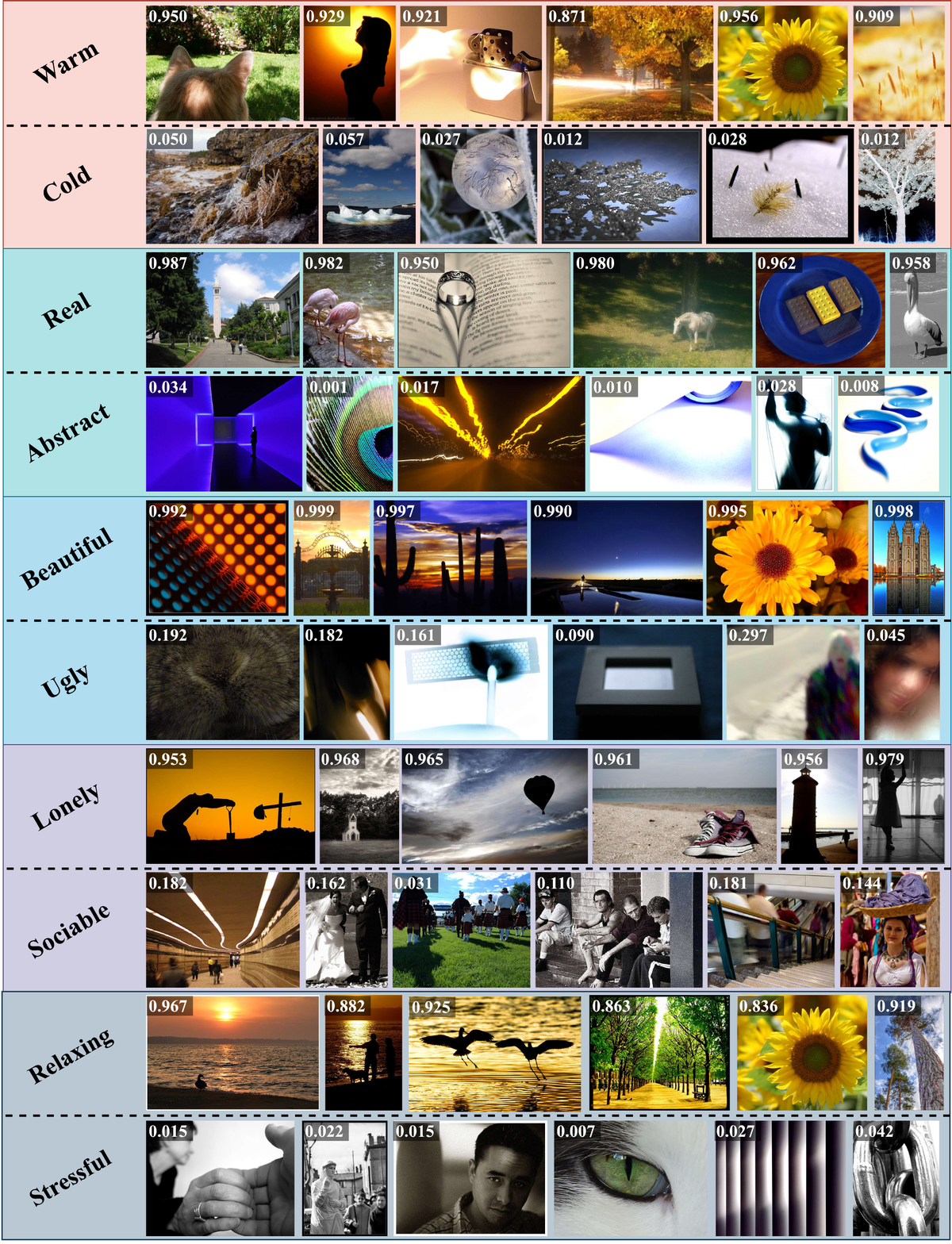}}
		\caption{\footnotesize{More examples of CLIP-IQA for assessing abstract perception AVA~\cite{murray2012ava}. The results on these five attributes show that CLIP-IQA is still able to identify these attributes. The numbers at the top left corner of each image represent CLIP-IQA scores. \textbf{(Zoom-in for best view)}
		}}
		\label{figure_ava_exa_clip}
	\end{center}
\end{figure*}

\begin{figure*}[h]
	\begin{center}
		\centerline{\includegraphics[width=2.0\columnwidth]{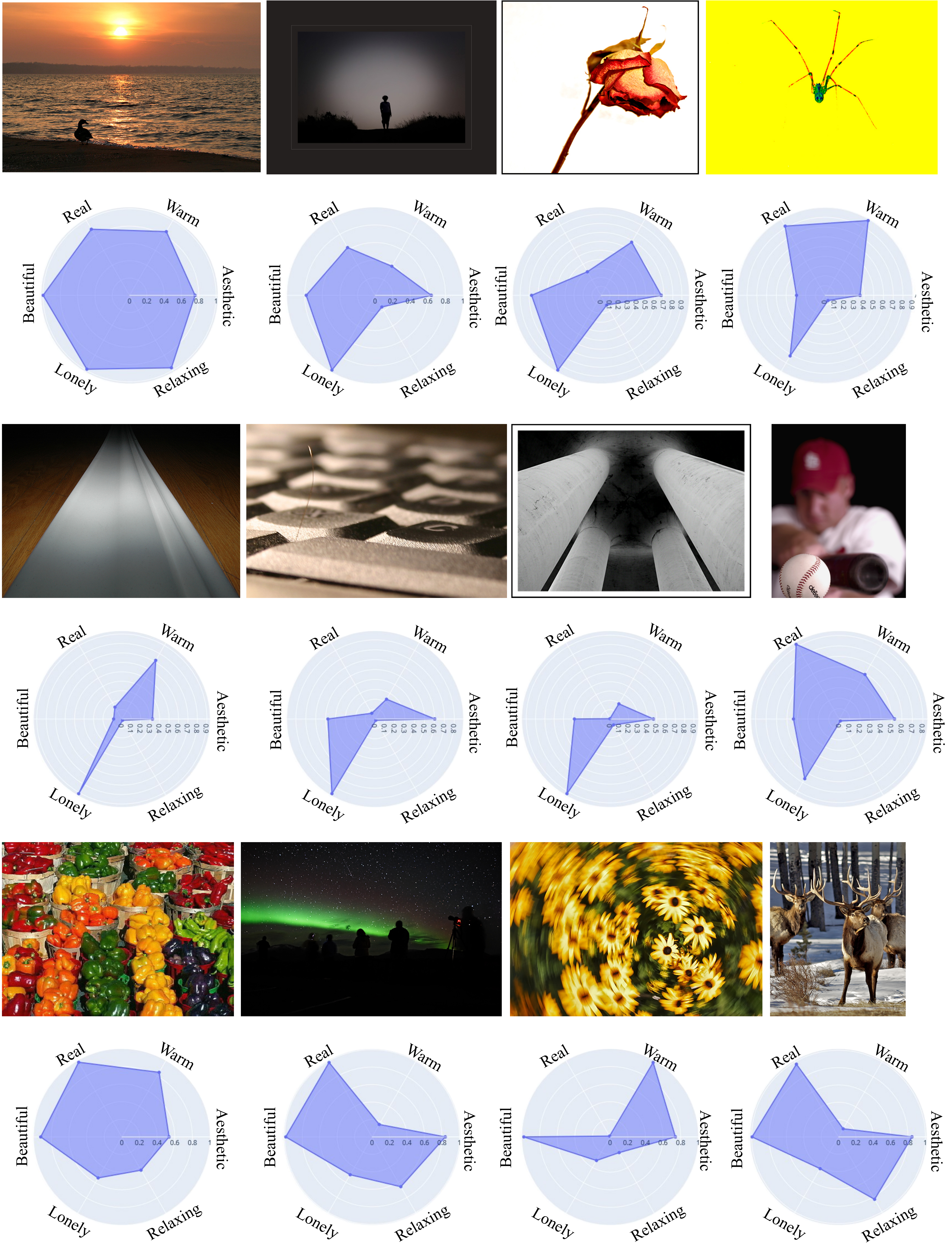}}
		\caption{\footnotesize{Examples of CLIP-IQA for assessing multiple abstract attributes on AVA~\cite{murray2012ava}. The results show that CLIP-IQA is able to identify these attributes. \textbf{(Zoom-in for best view)}
		}}
		\label{figure_ava_radar}
	\end{center}
\end{figure*}

In addition to the five abstract attributes shown in our main paper (\ie,~\textit{complex/simple}, \textit{natural/synthetic}, \textit{happy/sad}, \textit{scary/peaceful}, and \textit{new/old}), we further provide more quantitative results related to other attributes on AVA \cite{murray2012ava} benchmark. The prompts are shown in Table~\ref{table_prompt_ava}.
As shown in Fig.~\ref{figure_ava_exa_clip} and Fig.~\ref{figure_ava_radar}, the proposed CLIP-IQA can also well perceive the five attributes.

\clearpage

\clearpage
\small
\bibliography{aaai23}
\end{document}

%% file: section/1_introduction.tex
\begin{figure}[htb!]
	\begin{center}
		\centerline{\includegraphics[width=1.0\columnwidth]{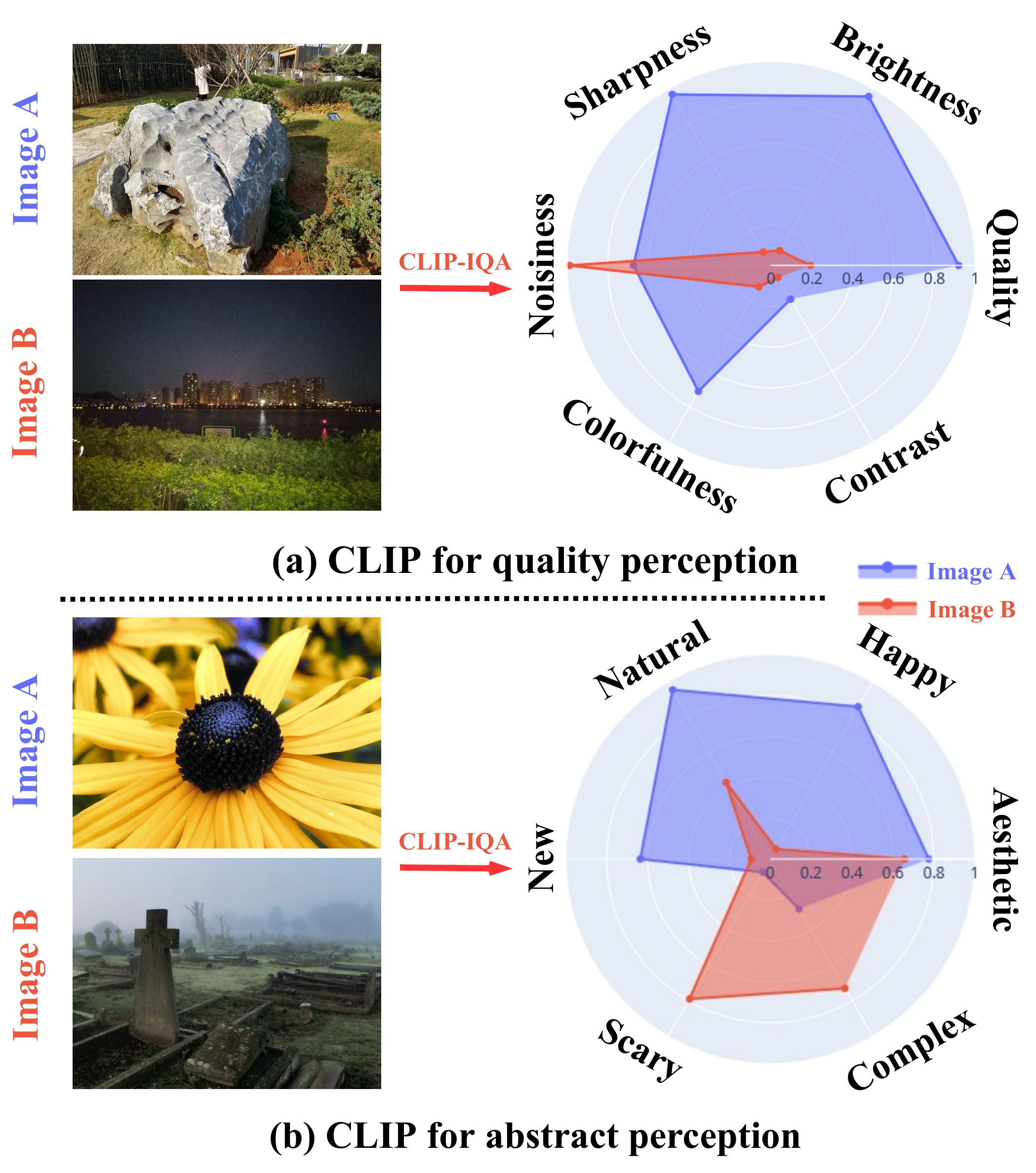}}
		\caption{We explore the potential of CLIP for visual perception. We find that CLIP is capable of assessing quality (\eg,~``overall quality'', ``brightness'', etc) as shown in (a). CLIP can be extended to the challenging task of abstract perception (\eg,~``aesthetic'', ``happy'', etc) in (b). The assessment is done without explicit task-specific training using the proposed antonym prompts to harness prior in CLIP. 
		}
		\label{figure_1}
	\end{center}
\end{figure}

\section{Introduction}

The \textit{look} and \textit{feel} are two contributing factors when humans interpret an image, and the understanding of these two elements has been a long-standing problem in computer vision. The \textit{look} of an image is often related to quantifiable attributes that directly affect the content delivery, such as exposure and noise level. In contrast, the \textit{feel} of an image is an abstract concept immaterial to the content and cannot be easily quantified, such as emotion and aesthetics. It is of common interest to explore the possibility of a universal understanding of both \textit{look} and \textit{feel}, as it can not only save efforts on manual labeling but also facilitate the development of vision tasks such as restoration~\cite{zhang2019ranksrgan,jo2020investigating,wenlong2021ranksrgan,hui2021learning}.

Considerable efforts have been devoted to the assessment of both the quality perception (\ie,~look) and the abstract perception (\ie,~feel) of images (We do not intend to define a new or complete taxonomy of visual perception. The notions of quality perception and abstract perception are used in this paper mainly to facilitate our analysis.). Earlier methods~\cite{mittal2012no,saad2012blind,mittal2012making,zhang2015feature,ma2017learning} design image-level features, such as natural scene statistics~\cite{ruderman1994statistics}, for quality assessment. Despite the success of these methods, the optimality of hand-crafted features is often in doubt, and the correlation to human perception is inferior in general.
To alleviate the problem, later methods~\cite{ma2017end,ke2021musiq,kong2016aesthetics,sheng2018attention,jin2022pseudo,kim2018building,yao2020apse,achlioptas2021artemis} resort to the learning-based paradigm, where a quality prediction model is directly learned from manually labeled data. With the presence of labeled data, these approaches demonstrate a much higher correlation to human perception. However, the laborious data labeling process and the task-specific nature of learning-based methods limit their generalizability to unseen attributes. More importantly, the aforementioned approaches are intended for either quality or abstract perception, possessing limited versatility\footnote{More discussion of related work is included in the supplementary material.}.

We are interested in knowing whether there exists a notion that (1) does not require hand-crafted features and task-specific training, (2) possesses a high correlation to human perception, and (3) can handle both quality and abstract perception. To this end, 
we turn our focus to CLIP \cite{radford2021learning}, a contrastive-learning-based visual-language pretraining approach.
Through training with massive image-text pairs, CLIP demonstrates exceptional capability in building semantic relationship between texts and visual entities without explicit training~\cite{patashnik2021styleclip,jain2021putting,hessel2021clipscore,rao2021denseclip,zhou2021denseclip,gu2021open}. 
Inspired by the appealing property of CLIP, we hypothesize that CLIP could have captured the relationship between human language and visual perception for image assessment. 

The problem of harnessing CLIP for perception assessment can be more challenging compared to existing works related to objective attributes, such as image manipulation \cite{patashnik2021styleclip,gabbay2021image,xu2021predict}, object detection \cite{gu2021open,zhong2021regionclip,shi2022proposalclip}, and semantic segmentation \cite{rao2021denseclip,zhou2021denseclip}.
Specifically, CLIP is known to be sensitive to the choices of prompts \cite{radford2021learning}, and perception is an abstract concept with no standardized adjectives, especially for the \textit{feel} of images. In addition, linguistic ambiguity~\cite{khurana2017natural} (\eg, ``a clean image'' can either refer to an image without noise or an image related to the action of cleaning) could severely affect perception assessment. As a result, the performance of CLIP on this task can be highly volatile, resulting from different choices of prompts.

Our study represents the first attempt to investigate the potential of CLIP on the challenging yet meaningful task of perception assessment. We begin our exploration by delving into the selection of prompts so that potential vagueness due to linguistic ambiguity can be minimized.
To this end, we introduce a prompt pairing
strategy where antonym prompts are adopted in pairs (\eg,~``Good photo.'' and ``Bad photo.'').
Based on our strategy, we show that CLIP can be directly applied to visual perception assessment without any task-specific finetuning.
%
For quality perception, we demonstrate that CLIP is able to assess the overall quality of an image by simply using ``good'' and ``bad'' as prompts, and achieve a high correlation to human's perception in common IQA datasets~\cite{ghadiyaram2015massive,hosu2020koniq,fang2020perceptual}.
Furthermore, we investigate the capability of CLIP in assessing fine-grained quality, such as brightness and noisiness. We apply CLIP on common restoration benchmarks \cite{Chen2018Retinex,fivek,xu2018real,rim_2020_ECCV} and synthetic data using fine-grained attributes, and it is shown that CLIP is capable of determining the fine-grained quality of an image (Fig.~\ref{figure_1}-(a)). 
In addition, the versatility of CLIP can be seen from its extension to abstract perception. In particular, CLIP also possesses superior performance when the quality-related prompts are replaced with abstract attributes (\eg,~``happy'' and ``sad''). Our results on both aesthetic benchmark \cite{murray2012ava} and corresponding user studies show that CLIP succeeds in distinguishing images with different \textit{feelings} following human perception (Fig.~\ref{figure_1}-(b)).

Given the success of CLIP in a wide range of vision tasks, we believe our exploration is timely. A powerful and versatile method for image assessment is indispensable. As the first work in this direction, we begin with a direct adaptation of CLIP to image assessment with a carefully designed prompt pairing strategy (named CLIP-IQA), followed by extensive experiments to examine the capability boundary of CLIP. We show that CLIP is able to assess not only the \textit{look} but also the \textit{feel} of an image to a satisfactory extent. We further analyze and discuss the limitations of CLIP on the task to inspire future studies.

%% file: section/3_methodology.tex
\begin{figure*}[htbp]
	\begin{center}
		\centerline{\includegraphics[width=2.0\columnwidth]{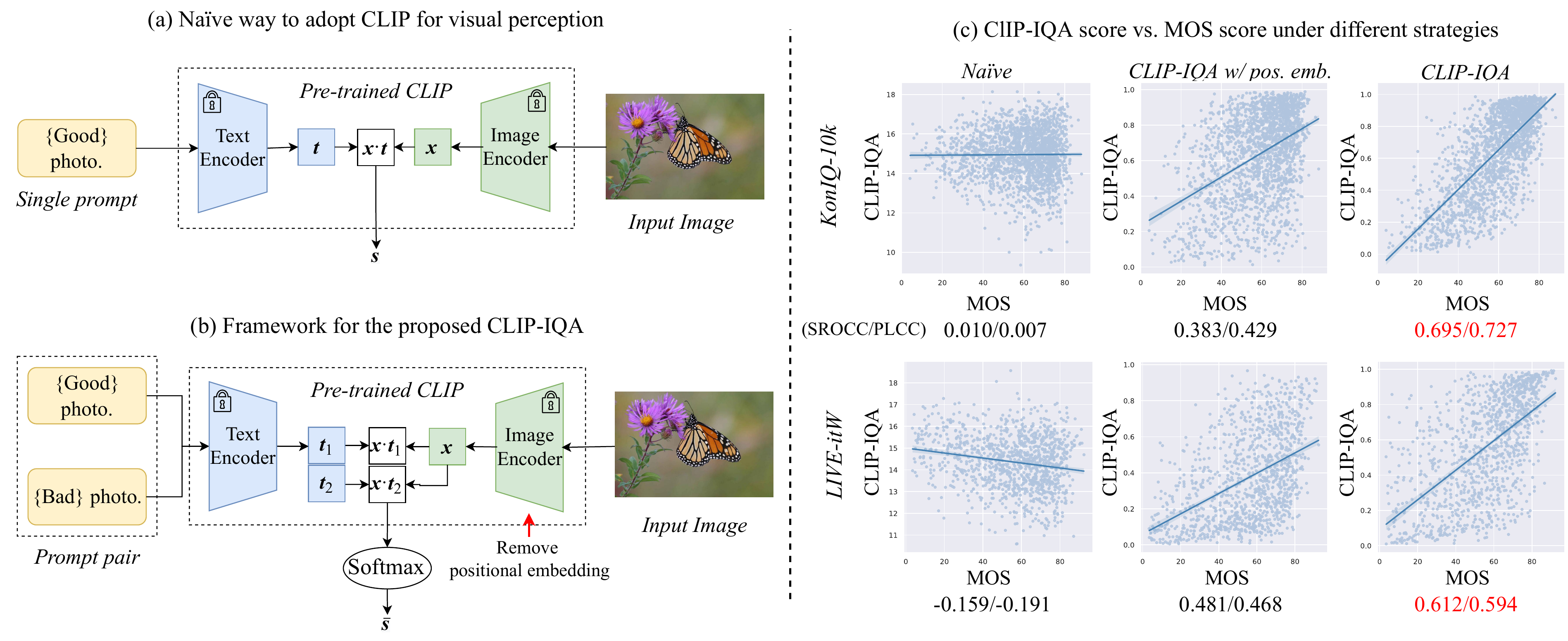}}
		\caption{(a) A na\"{i}ve approach of using a single prompt for perception assessment. (b) Our approach with (1) antonym prompt pairing strategy and (2) positional embedding removed. (c) Our proposed modifications obtain much higher correlations with human-labeled MOS score in both KonIQ-10k~\cite{hosu2020koniq} and LIVE-itW~\cite{ghadiyaram2015massive}. The numbers in (c) represent SROCC/PLCC (higher is better).
		}
		\label{figure_2}
	\end{center}
\end{figure*}

\section{CLIP for Visual Perception}
\label{Sec_3}

Exploiting CLIP as a versatile prior for visual perception assessment is not straightforward due to linguistic ambiguity inherent in this task. Next, we present our modifications to CLIP through prompt pairing (Sec.~\ref{CLIP_architecture}). We then examine the capability of CLIP for quality perception in Sec.~\ref{CLIP_quality}, followed by the extension to abstract perception in Sec.~\ref{CLIP_abstract}.

\subsection{Extending CLIP for Visual Perception}
\label{CLIP_architecture}

\textbf{Antonym prompt pairing.} As depicted in Fig.~\ref{figure_2}-(a), a straightforward way to exploit CLIP for perception assessment is to directly calculate the cosine similarity between the feature representations of a given prompt (\eg,~``Good photo'') and a given image. 
Specifically, let $\bm{x}\in\mathbb{R}^{C}$ and $\bm{t}\in\mathbb{R}^{C}$ be the features (in vector form) from the image and prompt, respectively, where $C$ is the number of channels. The final predicted score $s\in[0, 1]$ is computed as:
\begin{equation}
\label{eq:naive}
s = \dfrac{\bm{x}\odot\bm{t}}{||\bm{x}||\cdot||\bm{t}||},
\end{equation}
where $\odot$ denotes the vector dot-product and $||\cdot||$ represents the $\ell_2$ norm. 

While this na\"{i}ve adoption achieves a huge success in existing literature~\cite{patashnik2021styleclip,hessel2021clipscore,rao2021denseclip,zhou2021denseclip}, it is not viable for perception assessment, due to linguistic ambiguity~\cite{khurana2017natural}. 
Particularly, ``a rich image'' can either refer to an image with rich content or an image related to wealth.
As shown in Fig.~\ref{figure_2}-(c), using CLIP with a single prompt shows poor correlation with human perception on common IQA datasets~\cite{ghadiyaram2015massive,hosu2020koniq}.

To address this problem, we propose a simple yet effective prompt pairing strategy. In order to reduce ambiguity, we adopt antonym prompts (\eg,~``Good photo.'' and ``Bad photo.'') as a pair for each prediction. Let $\bm{t}_1$ and $\bm{t}_2$ be the features from the two prompts opposing in meanings, we first compute the cosine similarity
\begin{equation}
\label{eq:cosine}
s_i = \dfrac{\bm{x}\odot\bm{t}_i}{||\bm{x}||\cdot||\bm{t}_i||},\quad i \in \{1,2\},
\end{equation}
and Softmax is used to compute the final score $\bar{s}\in[0, 1]$:
\begin{equation}
\label{eq:pair}
\bar{s} = \frac{e^{s_1}}{e^{s_1} + e^{s_2}}.
\end{equation}
When a pair of adjectives is used, the ambiguity of one prompt is reduced by its antonym as the task is now cast as a binary classification, where the final score is regarded as a relative similarity. In particular, a larger $\bar{s}$ indicates a closer match to the corresponding attribute of $\bm{t}_1$.
As shown in Fig.~\ref{figure_2}-(c), our proposed antonym prompt pairing significantly boosts the performance of CLIP -- our method can predict scores more consistent with human-labeled MOS scores, reflecting from the higher Spearman's Rank-order Correlation Coefficient (SROCC) and Pearson's Linear Correlation Coefficient (PLCC).

\noindent\textbf{Removal of positional embedding.}
Another limitation of CLIP is the requirement of fix-sized inputs. In particular, the ResNet-50-based CLIP requires an image with a size of $224{\times}224$. Such a requirement is unfavorable in perception assessment as the resizing and cropping operations may introduce additional distortions to the input image, altering the final score. For instance, the resizing operation leads to insensitivity to image resolution, and hence assessments related to resolution could be futile.

The above limitation results from the learnable positional embedding, which is fixed in size upon construction. Different from existing studies~\cite{patashnik2021styleclip,hessel2021clipscore,jain2021putting,gu2021open} that extensively use positional embedding, we conjecture that the positional embedding has a minimal effect on perception assessment as the primary focus of this task is to capture the perceptual relationship between an image and a given description. Therefore, we propose to remove the positional embedding to relax the size constraint. 
We adopt the ResNet variant since it can provide a stronger inductive bias to complement the removal of positional information. The comparison between different backbones is discussed in Sec.~\ref{sec_net}.
As shown in Fig.~\ref{figure_2}-(c), this relaxation further improves the correlation with human perception. We term our model \textbf{CLIP-IQA}, and all our experiments follow this setting.

\begin{figure*}[t]
	\begin{center}
		\centerline{\includegraphics[width=2.0\columnwidth]{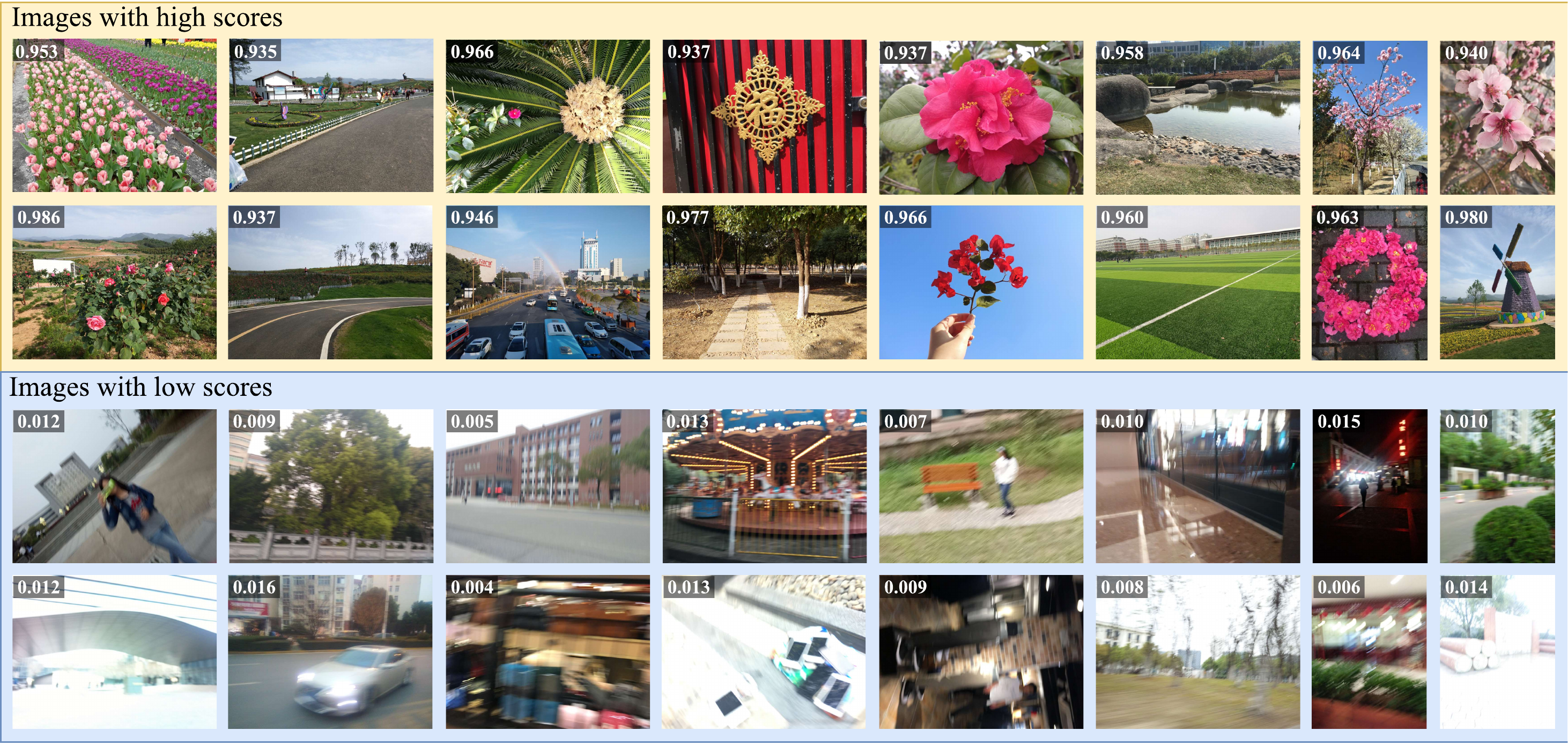}}
		\caption{Best-scored and worst-scored images labeled by CLIP-IQA in SPAQ~\cite{fang2020perceptual}. CLIP-IQA performs well on SPAQ, i.e., assigning high scores to images with clear details and low scores to images with blur and low contrast. The numbers at the top left corner of each image indicate the CLIP-IQA score. \textbf{(Zoom-in for best view)}}
		\label{figure_iqa}
	\end{center}
\end{figure*}

\subsection{Quality Perception}
\label{CLIP_quality}
This section focuses on exploring the potential of CLIP-IQA on quality perception assessment. Specifically, we investigate its effectiveness on assessing the overall quality of images using conventional No-Reference IQA (NR-IQA) datasets~\cite{ghadiyaram2015massive,hosu2020koniq,fang2020perceptual}. We then extend our scope to fine-grained attributes using synthetic data and common restoration benchmarks~\cite{Chen2018Retinex,fivek,xu2018real,rim_2020_ECCV}.

\begin{table}[t]
    \newcommand{\tabincell}[2]{\begin{tabular}{@{}#1@{}}#2\end{tabular}}
    \renewcommand\arraystretch{1.2}
	\begin{center}
		\caption{Comparison to NR-IQA methods -- Type A: Methods w/o task-specific training, Type B: Learning-based methods w/ task-specific training (trained on KonIQ-10k~\cite{hosu2020koniq} and tested on KonIQ-10k, LIVE-itW~\cite{ghadiyaram2015massive} and SPAQ~\cite{fang2020perceptual}). \rf{Red} and \bd{blue} colors represent the best and second best performance, respectively.}
		\label{table_iqa}
		\resizebox{0.48\textwidth}{!}{
		\begin{tabular}{c|c|cc|cc|cc}
            \toprule \multirow{2}{*}{Type} & \multirow{2}{*}{Methods} &  \multicolumn{2}{c|}{KonIQ-10k} & \multicolumn{2}{c|}{LIVE-itW} & \multicolumn{2}{c}{SPAQ}\\
            \cline{3-8} & & SROCC\,$\uparrow$ & PLCC\,$\uparrow$ & SROCC\,$\uparrow$ & PLCC\,$\uparrow$ & SROCC\,$\uparrow$ & PLCC\,$\uparrow$ \\
            \hline \multirow{5}{*}{\makecell[c]{A}} & BIQI & 0.559 & 0.616 & 0.364 & 0.447 & 0.591 & 0.549 \\ 
            \cline{2-8} & BLIINDS-II & 0.585 & 0.598  & 0.090 & 0.107 & 0.317 & 0.326 \\
            \cline{2-8} & BRISQUE & \rf{0.705} & \bd{0.707} & \bd{0.561} & \rf{0.598} & 0.484 & 0.481 \\
            \cline{2-8} & NIQE & 0.551 & 0.488 & 0.463 & 0.491 & \bd{0.703} & \bd{0.670} \\
            \cline{2-8} & \textbf{CLIP-IQA} & \bd{0.695} & \rf{0.727} & \rf{0.612} & \bd{0.594} & \rf{0.738} & \rf{0.735} \\
            \hline \multirow{6}{*}{\makecell[c]{B}} & CNNIQA & 0.572 & 0.584 & 0.465 & 0.450 & 0.664 & 0.664 \\
            \cline{2-8} & KonCept512 & \bd{0.921} & \rf{0.937} & \rf{0.825} & \rf{0.848} & 0.837 & 0.815 \\
            \cline{2-8} & HyperIQA & 0.904 & 0.915 & 0.760 & 0.776 & 0.811 & 0.805 \\
            \cline{2-8} & MUSIQ & \rf{0.924} & \rf{0.937} & 0.793 & \bd{0.832} & \rf{0.873} & \rf{0.868} \\
            \cline{2-8} & VCRNet & 0.894 & 0.909 & 0.678 & 0.701 & 0.781 & 0.766 \\
            \cline{2-8} & \textbf{CLIP-IQA$^+$} & 0.895 & 0.909 & \bd{0.805} & \bd{0.832} & \bd{0.864} & \bd{0.866} \\
            \bottomrule
		\end{tabular}
}
	\end{center}
\end{table}

\begin{figure}[t]
	\begin{center}
		\centerline{\includegraphics[width=1.0\columnwidth]{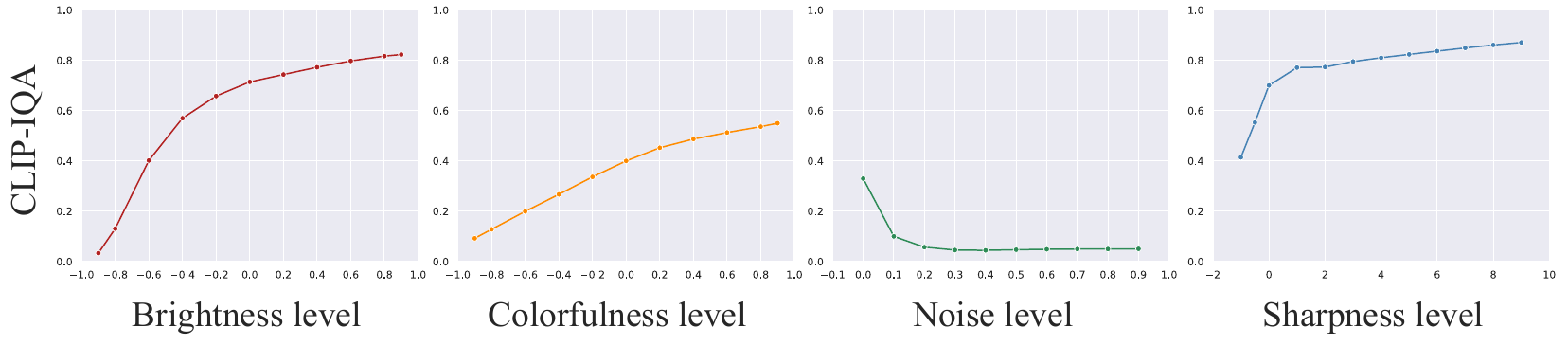}}
		\caption{CLIP-IQA for fine-grained attributes on synthetic data with different input scales. CLIP-IQA clearly demonstrates positive correlations to the change of attributes.}
		\label{figure_synthetic}
	\end{center}
\end{figure}

\begin{figure*}[tbp]
	\begin{center}
		\centerline{\includegraphics[width=2.0\columnwidth]{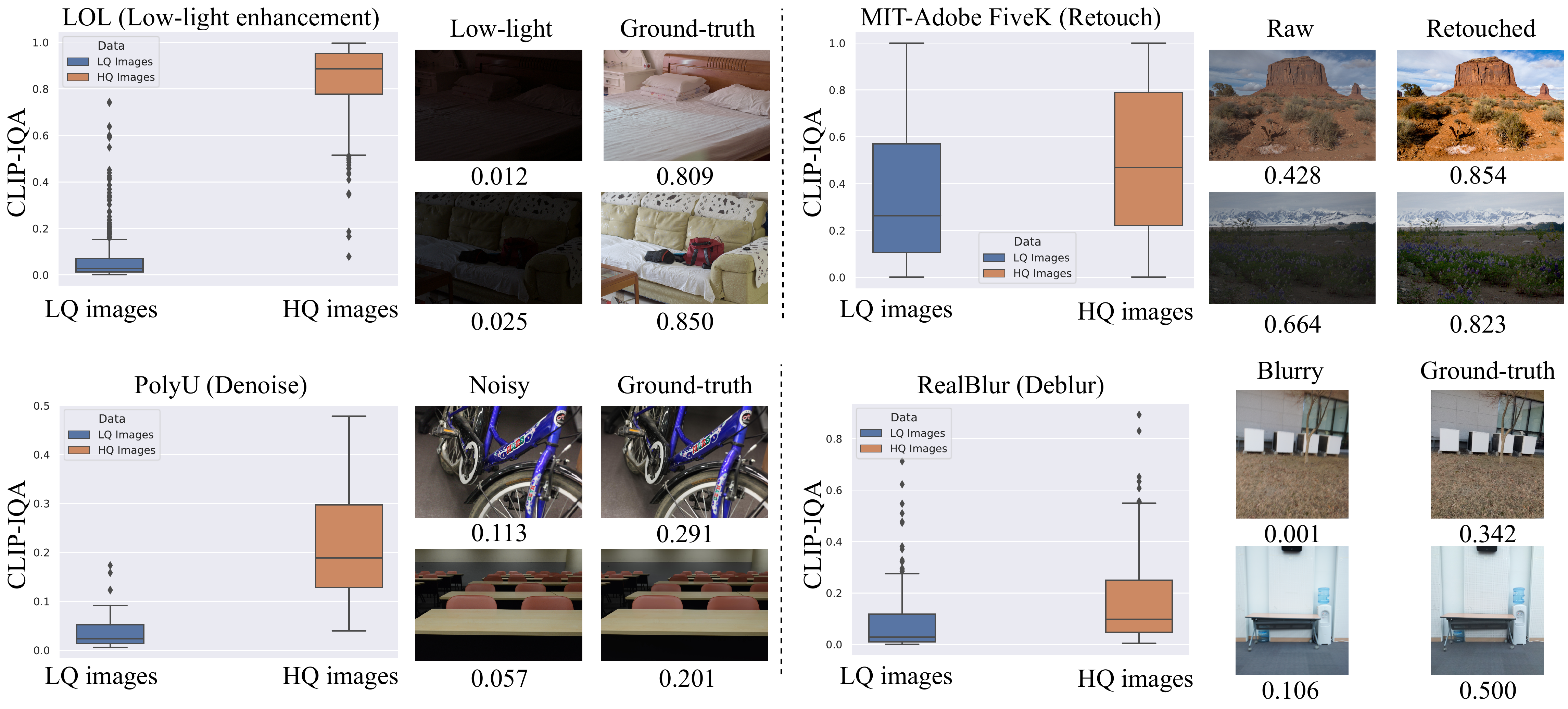}}
		\caption{CLIP-IQA for fine-grained attributes on real-world benchmarks. Low-quality images receive a lower score from CLIP-IQA. The numbers under each image are the corresponding score from CLIP-IQA.}
		\label{figure_box}
	\end{center}
\end{figure*}

\noindent\textbf{CLIP-IQA for overall quality.} 
To assess the overall quality, we simply use one of the most commonly seen antonym 
\begin{center}
    [``\texttt{Good photo.}'', ``\texttt{Bad photo.}'']    
\end{center} 
as the paired prompt for CLIP-IQA. We conduct experiments on three widely used NR-IQA benchmarks including LIVE-itW~\cite{ghadiyaram2015massive} and KonIQ-10k~\cite{hosu2020koniq} for realistic camera distortions and SPAQ~\cite{fang2020perceptual} for smartphone photography.
We compare CLIP-IQA with nine NR-IQA methods, including four non-learning-based methods: BIQI~\cite{moorthy2010two}, BLIINDS-II \cite{saad2012blind}, BRISQUE~\cite{mittal2012no}, NIQE~\cite{mittal2012making}, and five learning-based methods: CNNIQA~\cite{kang2014convolutional}, KonCept512~\cite{hosu2020koniq}, HyperIQA~\cite{su2020blindly}, MUSIQ~\cite{ke2021musiq}, VCRNet~\cite{pan2022vcrnet}.
For the non-learning-based methods, we take the numbers from \cite{hosu2020koniq} for LIVE-itW and KonIQ-10k and adopt the official code
for SPAQ. For learning-based methods except for MUSIQ\footnote{Since training code is unavailable, we adopt the official model trained on KonIQ-10k.}, following  Hosu~\etal~\cite{hosu2020koniq}, we train the models on KonIQ-10k dataset and test on the three datasets to assess their generalizability.
Performance is evaluated with SROCC and PLCC. Detailed settings are discussed in the supplementary material.

As shown in Table~\ref{table_iqa}, without the need of hand-crafted features, CLIP-IQA is comparable to BRISQUE and surpasses all other non-learning methods on all three benchmarks. In addition, even without task-specific training, CLIP-IQA outperforms CNNIQA, which requires training with annotations. The surprisingly good performance of CLIP-IQA verifies its potential in the task of NR-IQA. In Fig.~\ref{figure_iqa} we show two sets of images obtaining high scores and low scores from CLIP-IQA respectively. It is observed that CLIP-IQA is able to distinguish images with different qualities.

In scenarios where annotations are available, CLIP-IQA can also be finetuned for enhanced performance. In this work, we develop a simple extension of CLIP-IQA, named CLIP-IQA$^+$, by using CoOp~\cite{zhou2021learning} to finetune the prompts. Specifically, the prompts are initialized as [``\texttt{Good photo.}'', ``\texttt{Bad photo.}''] and updated with standard backpropagation\footnote{We adopt SGD with a learning rate of 0.002 during training. The model is trained for 100000 iterations with batch size 64 on KonIQ-10k dataset and the MSE loss is used to measure the distance between the predictions and the labels.}. The network weights are kept fixed.
From Table~\ref{table_iqa} we see that CLIP-IQA$^+$ achieves SROCC/PLCC of 0.895/0.909, 0.805/0.832 and 0.864/0.866 on KonIQ-10k, LIVE-itW and SPAQ, respectively. These results are comparable to state-of-the-art deep learning based methods \cite{hosu2020koniq,su2020blindly,ke2021musiq,pan2022vcrnet}, showing the potential of large-scale language-vision training. Furthermore, it is observed that CLIP-IQA$^+$ possesses better generalizability. Specifically, while most learning-based methods experience significant performance drops on LIVE-itW and SPAQ, the drop of CLIP-IQA$^+$ is more subtle, and CLIP-IQA$^+$ achieves the second-best performance without task-specific architecture designs. We believe that the performance of CLIP-IQA$^+$ can be further improved by replacing CoOp with more sophisticated finetuning schemes such as Tip-Adaptor~\cite{zhang2021tip}.
In addition to the generalizability, another advantage of CLIP-IQA$^+$ is the low storage cost for domain transfer. Unlike existing NR-IQA methods (\eg, MUSIQ) that train different models for different domains (\eg, technical quality, aesthetics quality), CLIP-IQA$^+$ needs to store only two prompts for each domain, significantly reducing the storage cost.

\begin{figure*}[tp]
	\begin{center}
		\centerline{\includegraphics[width=2.0\columnwidth]{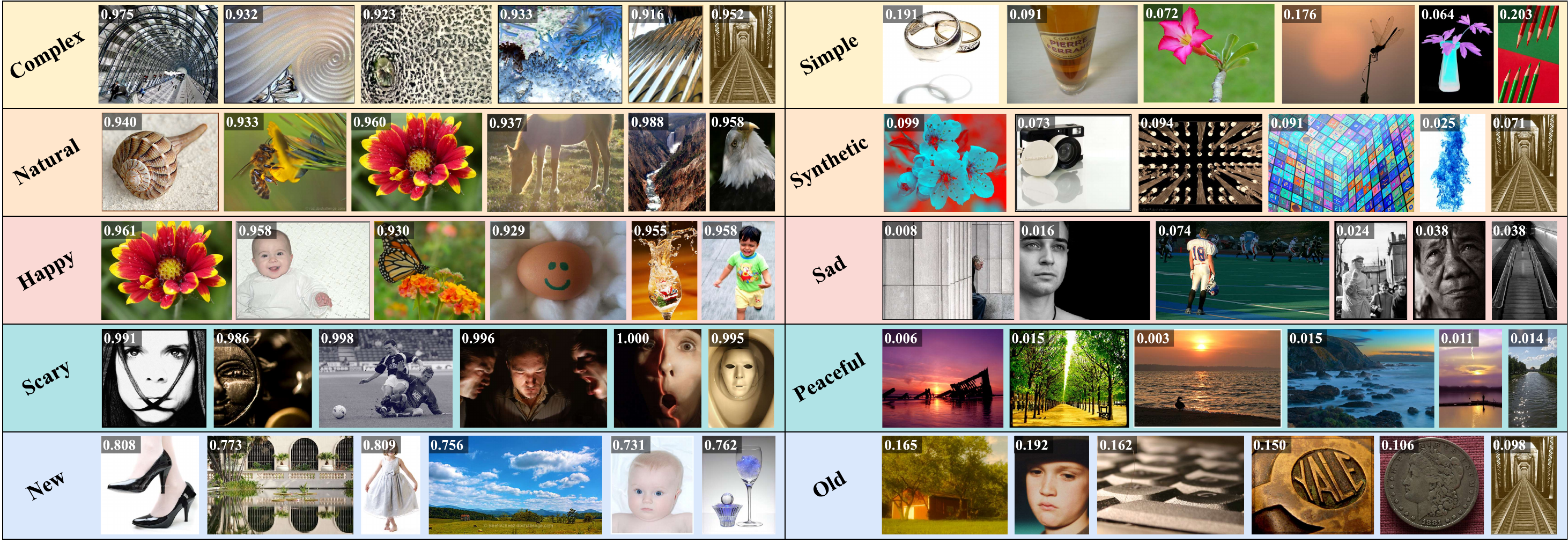}}
		\caption{CLIP-IQA for assessing abstract perception. The results on these five attributes show that CLIP-IQA is able to understand abstract perception, \eg~\textit{``Complex/Simple''}. The left and right are two sets of images selected according to the CLIP-IQA scores (numbers at the top left corner of each image) on abstract-attribute antonym pairs. \textbf{(Zoom-in for best view)}}
		\label{figure_abstract}
	\end{center}
\end{figure*}

\begin{figure}[h]
	\begin{center}
		\centerline{\includegraphics[width=1.0\columnwidth]{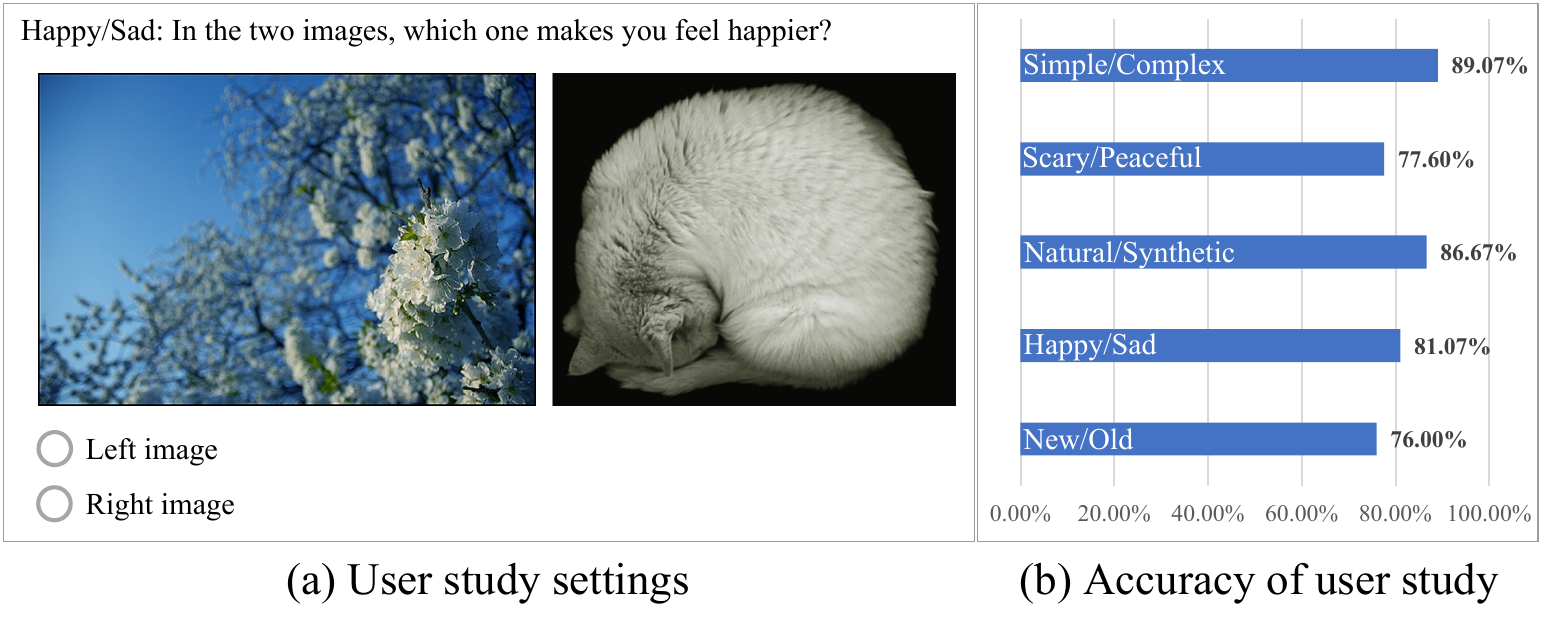}}
		\caption{User study for abstract perception. (a) We ask the subjects to compare the two images and select the one more consistent with the given description. (b) The prediction of CLIP-IQA is consistent with human about 80\% of the time.}
		\label{figure_mos}
	\end{center}
\end{figure}

\noindent\textbf{CLIP-IQA for fine-grained quality.} 
Instead of assigning a single score to an image, humans usually judge an image from multiple perspectives, such as brightness, noisiness, and sharpness. Given the success of CLIP-IQA on the overall quality, we are curious about whether it can go beyond a single score and assess an image in a fine-grained fashion. To test the capability, we simply replace ``good'' and ``bad'' with the attribute we want to assess and its antonym, respectively. For example, we use ``\texttt{Bright photo.}'' and ``\texttt{Dark photo.}'' as the prompt when evaluating the brightness of images. It is noteworthy that unlike most learning-based approaches, CLIP-IQA does not require quality annotations. Therefore, CLIP-IQA is not confined to annotated labels and is able to assess arbitrary attributes required by users.

We first conduct our experiments on synthetic data for four representative attributes: \textit{brightness}, \textit{noisiness}, \textit{colorfulness}, and \textit{sharpness}. We adjust the extent of each attribute using the Python package \texttt{PIL}, and evaluate the performance on 200 images randomly selected from the test set of KonIQ-10k. From Fig.~\ref{figure_synthetic} we see that CLIP-IQA exhibits a high correlation to the changes of the attributes. More details are included in the supplementary material.

To further confirm its real-world applicability, we apply CLIP-IQA on four non-synthetic benchmarks: \textit{LOL}~\cite{Chen2018Retinex} for low-light enhancement, \textit{PolyU}~\cite{xu2018real} for denoising, \textit{MIT-Adobe FiveK}~\cite{fivek} for retouching, and \textit{RealBlur}~\cite{rim_2020_ECCV} for deblurring. For each dataset, we apply CLIP-IQA to the low-quality images and high-quality images (\ie,~ground-truths) independently, and compare their quality score. As depicted in Fig.~\ref{figure_box}, the quality scores of high-quality images are clearly above the low-quality ones, indicating that CLIP-IQA is able to identify fine-grained qualities.

\subsection{Abstract Perception}
\label{CLIP_abstract}
While emotions and aesthetics are readily comprehensible by humans, the understanding of such concepts remains non-trivial for machines. This section investigates the competence of CLIP-IQA in interpreting \textit{abstract perception}.
We conduct our experiments on the AVA dataset~\cite{murray2012ava} since it is among the largest visual perception datasets, containing over 250,000 images with a broad variety of content.

We evaluate the performance of CLIP-IQA using five attributes related to human emotions and artistic sensations, \ie, \textit{complex/simple}, \textit{natural/synthetic}, \textit{happy/sad}, \textit{scary/peaceful}, and \textit{new/old}. For each attribute, we compute the score for each image, and images are sorted according to their scores. In Fig.~\ref{figure_abstract} we show the best-scored and worst-scored images for each attribute. It is observed that CLIP-IQA is able to perceive the abstract aspects of an image. Similar to quality attributes, CLIP-IQA is able to assess an image from different perspectives. For instance, the last image in the \textit{complex} category is also \textit{synthetic} and \textit{old}. Such a fine-grained abstract understanding again demonstrates the effectiveness of the prior captured by CLIP during vision-language pretraining.

We further conduct a user study to verify the effectiveness of CLIP-IQA on abstract perception. The settings of the study are depicted in Fig.~\ref{figure_mos}-(a). Specifically, we generate 15 image pairs for each of the above five attributes. For each pair, we label the one with higher score as positive (\eg,~happy) and lower score as negative (\eg,~sad). 
Each pair is rated by 25 subjects. Subjects are asked to compare the two images and choose the one more consistent with the provided description (\eg,~In the two images, which one makes you feel happier?). We then compute the classification accuracy of CLIP-IQA using the subject scores as ground-truths.
As depicted in Fig.~\ref{figure_mos}, CLIP-IQA achieves an accuracy of about 80\% on all five attributes, indicating the great potential of CLIP-IQA on abstract idea understanding.

%% file: section/4_discussion.tex
\section{Discussion}
As covered in Sec.~\ref{CLIP_architecture}, the adaptation of CLIP to perception assessment requires careful designs on the prompts and backbones. In this section, we will delve deeper into the studies of the two components. Besides, we will discuss the limitations of CLIP-IQA to provide insights for future work.

\subsection{Prompt Designs} 
\label{sec_prompt}
While it is verified in previous works~\cite{radford2021learning,zhou2021learning,rao2021denseclip} that the selection of prompts could impact the performance, such conclusion has not been drawn in the task of perception assessment. Therefore, we conduct empirical studies to investigate the effects brought by different choices of prompts.

First, it is observed that the choice of prompt templates has a significant effect to CLIP-IQA. We compare the performance of three different choices of templates from existing literature~\cite{rao2021denseclip,zhou2021denseclip} -- (1)~``\texttt{[text] photo.}'', (2)~``\texttt{A photo of [text].}'', and (3)~``\texttt{There is [text] in the photo.}''. As shown in Table~\ref{table_baseline}, noticeable differences are observed when using different templates. In this work, we adopt ``\texttt{[text] photo.}'' for simplicity, and we believe that there exist more sophisticated templates that lead to enhanced performance. Such exploration is left as our future work.

Next, we investigate the influence of adjectives with the above template. Similarly, the performance varies with the adjectives selected\footnote{We collect adjectives from various image assessment benchmarks and photo sharing websites, which demonstrate decent results on KonIQ-10k and LIVE-itW.}. For instance, as shown in Table~\ref{table_baseline}, when assessing the overall quality of an image, ``\texttt{Good/Bad}'' achieves a better correlation to human perception than ``\texttt{High quality/Low quality}'' and ``\texttt{High definition/Low definition}''. We conjecture that such a phenomenon indicates poorer performance of uncommon adjectives. We remark here that this challenge is especially notable in perception assessment due to the existence of synonyms. The sensitivity to the choices of prompts indicates the need of a more comprehensive understanding of prompt designs.

\begin{table}[tp]
    \newcommand{\tabincell}[2]{\begin{tabular}{@{}#1@{}}#2\end{tabular}}
    \renewcommand\arraystretch{1.2}
	\begin{center}
		\caption{Comparison of CLIP-IQA variants. It is observed that the choices of prompts and backbones have a significant effect on the performance. The templates are (1)~``\texttt{[text] photo.}'', (2)~``\texttt{A photo of [text].}'', and (3)~``\texttt{There is [text] in the photo}, respectively. The adjectives are (a) ``\texttt{Good/Bad}'', (b) ``\texttt{High quality/Low quality}'', and (c) ``\texttt{High definition/Low definition}'', respectively. The first row is the settings for CLIP-IQA.}
		\label{table_baseline}
		\resizebox{0.48\textwidth}{!}{
		\begin{tabular}{c|c|c|c|cc|cc}
            \toprule \multicolumn{4}{c|}{Settings} & \multicolumn{2}{c|}{KonIQ-10k} & \multicolumn{2}{c}{LIVE-itW} \\
            \hline Template & Adjective & Backbone & Pos. Embedding & SROCC & PLCC & SROCC & PLCC \\
             \hline (1) & \multirow{3}{*}{\makecell[c]{(a)}} & ResNet-50 & \xmark & 0.695 & 0.727 & 0.612 & 0.594 \\
            \cline{1-1}  \cline{3-8} (2) & & ResNet-50 & \xmark & 0.116 & 0.119 & 0.263 & 0.276 \\
            \cline{1-1}  \cline{3-8} (3) & & ResNet-50 & \xmark & 0.214 & 0.217 & 0.347 & 0.351 \\
            \hline \multirow{2}{*}{\makecell[c]{(1)}} & (b) & ResNet-50 & \xmark & 0.537 & 0.570 & 0.462 & 0.429 \\
            \cline{2-8} &  (c) & ResNet-50 & \xmark & 0.592 & 0.580 & 0.611 & 0.560 \\
            \hline (1) & (a) & ResNet-50 & Vanilla & 0.383 & 0.429 & 0.481 & 0.400 \\
            \hline (1) & (a) & ResNet-50 & Interpolated & 0.682 & 0.690 & 0.583 & 0.555 \\
             \hline (1) & (a) & VIT-B/32 & Vanilla & 0.416 & 0.464 & 0.488 & 0.479 \\
            \hline (1) & (a) & VIT-B/32 & Interpolated & 0.634 & 0.643 & 0.503 & 0.491 \\
            \hline (1) & (a) &  VIT-B/32 & \xmark & 0.391 & 0.374 & 0.375 & 0.365 \\
            \bottomrule
		\end{tabular}
}
	\end{center}
\end{table}

\subsection{Backbone of Image Encoder}
\label{sec_net}
The backbone of the image encoder significantly affects the performance of CLIP-IQA. 
Unlike convolutional models that implicitly capture positional information~\cite{xu2021positional}, Transformer relies heavily on the positional embedding to provide such clue. Hence, the Transformer variant of CLIP incorporates the positional embedding in earlier layers to compensate for the lack of such inductive bias. In contrast, the ResNet variant depends less on the positional embedding, and adopts the embedding only in deep layers during multi-head attention. Therefore, it is expected that the Transformer variant would experience a larger performance drop when the positional embedding is removed.

To verify the hypothesis, we conduct experiments on KonIQ-10k and LIVE-itW.
As shown in Table~\ref{table_baseline}, while the Transformer variant shows better performance than the ResNet variant with the positional embedding, a significant drop is observed in the Transformer variant when the embedding is removed. 
This result corroborates our hypothesis that positional embedding is more crucial in Transformer. 

We further compare the performance of the ResNet variants with (1) positional embedding removal, and (2) positional embedding interpolation. We notice from Table~\ref{table_baseline} that the performance of the ResNet variant with positional embedding removal is better than that using positional embedding interpolation. We conjecture this is due to the inaccurate positional embedding caused by interpolation. Given the importance of accepting arbitrary-sized inputs in perception assessment, we adopt the ResNet variant with positional embedding removed throughout our experiments.

\subsection{Limitations}
\label{sec_limit}
Despite the encouraging performance of CLIP-IQA, there are challenges that remain unresolved. First, as discussed in Sec.~\ref{sec_prompt}, CLIP-IQA is sensitive to the choices of prompts. Therefore, it is of great importance to develop more systematic selection of prompts. In the future, our attention will be devoted to the design of prompts for improved performance. 

Second, through training with vision-language pairs, CLIP is able to comprehend words and phrases widely used in daily communication. However, it remains a non-trivial problem for CLIP to recognize professional terms that are relatively uncommon in human conversations, such as \textit{``Long exposure''}, \textit{``Rule of thirds''}, \textit{``Shallow DOF''}. Nevertheless, we believe that this problem can be attenuated by pretraining CLIP with such pairs.

Third, although our exploration shows the capability of CLIP on versatile visual perception tasks without explicit task-specific training, there still exist performance gaps between CLIP-IQA and existing task-specific methods due to the lack of task-specific architecture designs in CLIP-IQA. We believe the fusion of task-specific designs and the vision-language prior would achieve promising performance and will be a new direction in the task of perception assessment.

%% file: section/5_conclusion.tex
\section{Conclusion}
The remarkable versatility of CLIP has aroused great interests from computer vision researchers. This paper diverges from existing works and investigates the effectiveness of CLIP on perceiving subjective attributes. From our exploration, we find that CLIP, when equipped with suitable modifications, is able to understand both quality and abstract perceptions of an image.
By providing a solid ground for CLIP-based perception assessment, we believe our studies and discussion could motivate future development in various domains, such as sophisticated prompts, better generalizability, and effective adoption of CLIP prior. 

%% file: section/2_relatedworks.tex
\section{Related Work}

\textbf{Quality perception.}
Quality perception of an image mainly focuses on quantifiable attributes such as exposure and noise level.
Existing methods mainly focus on assessing the overall quality of an image. Image Quality Assessment (IQA) can be mainly classified into Full-Reference IQA (FR-IQA)~\cite{video2000final,wang2003multiscale,wang2004image,sheikh2006image,zhang2011fsim,xue2013gradient,zhang2014vsi,blau2018perception,zhang2018unreasonable,prashnani2018pieapp,cheon2021perceptual,ding2021comparison,jinjin2020pipal} and No-Reference IQA (NR-IQA)~\cite{ye2012unsupervised,mittal2012no,saad2012blind,mittal2012making,kang2014convolutional,zhang2015feature,ma2017learning,bosse2017deep,ma2017end,zhang2018blind,su2020blindly,zhu2020metaiqa,ke2021musiq,pan2022vcrnet}.
The objective of FR-IQA is to quantify the similarity between two images, and it has been widely adopted in evaluating the quality of image coding~\cite{kopilovic2005artifact,wang2011applications} and restoration~\cite{jo2020investigating,wenlong2021ranksrgan}.
PSNR~\cite{video2000final} and SSIM~\cite{wang2004image} are among the most common metrics in the FR-IQA paradigm, thanks to their simplicity. However, it is well-known~\cite{blau2018perception,zhang2018unreasonable} that such metrics have a weak correlation to human perception. Therefore, a prevalent direction is to develop full-reference metrics that correlate well with humans~\cite{zhang2018unreasonable,prashnani2018pieapp,jinjin2020pipal,cheon2021perceptual}. For instance, LPIPS~\cite{zhang2018unreasonable} shows a better correlation when computing the distance in the feature domain, using a pretrained network.

The main limitation of FR-IQA is that it is confined to scenarios where references exist. In order to assess arbitrary images in the wild, NR-IQA aims at evaluating image quality without the use of references.
Conventional NR-IQA methods \cite{mittal2012no,saad2012blind,mittal2012making,zhang2015feature,ma2017learning} mainly rely on natural scene statistics (NSS) \cite{ruderman1994statistics}, a model revealing statistic property of natural images. The core idea of these methods is to hand-craft degradation-aware features to characterize NSS and obtain the final quality score using regression models \cite{mittal2012no,ma2017learning} or multivariate Gaussian model \cite{saad2012blind,mittal2012making,zhang2015feature}.
Recently, benefiting from the rapid development of deep learning, many learning-based methods \cite{kang2014convolutional,bosse2017deep,ma2017end,zhang2018blind,su2020blindly,zhu2020metaiqa,ke2021musiq,pan2022vcrnet} have been proposed and shown superior performance compared with conventional methods.
However, these methods requires training with human-labeled benchmarks \cite{larson2010most,sheikh2006statistical,ponomarenko2015image,ghadiyaram2015massive,hosu2020koniq,fang2020perceptual}, limiting their generalizability and versatility.
In this work, we devote our attention to a more general task of \textit{perception assessment}. Specifically, we focus on not only overall quality of an image, but also the fine-grained \textit{quality attributes} as well as \textit{abstract attributes}. The need of a versatile and generalizable assessment motivates the use of CLIP, which is trained with large-scale image-text pairs. Surprisingly, we find that our CLIP-IQA achieves comparable performance to existing IQA methods, and can be extended to assessing various aspects of an image.

\noindent\textbf{Abstract perception.}
Abstract perception is a subjective feeling and is not quantifiable.
Existing methods for abstract perception assessment mainly focus on aesthetic image assessment \cite{murray2012ava,kong2016aesthetics,sheng2018attention,zeng2019unified,shu2021semi,jin2022pseudo} and image emotion analysis \cite{calvo2004gaze,zhang2013affective,he2015image,kim2018building,panda2018contemplating,yao2020apse,achlioptas2021artemis}.
The majority of these methods require human annotations for training, which are laborious to obtain. Therefore, in this work, we explore the possibility of exploiting vision-language priors captured in CLIP to bypass the quality labeling process. Our results show that our CLIP-IQA is able to perceive abstract aspects of an image, even without training with annotations.

\noindent\textbf{CLIP-based approaches.}
Benefiting from the large-scale visual-language training, CLIP has shown impressive capability and generalizability on a wide range of tasks, such as text-driven image manipulation \cite{patashnik2021styleclip,gabbay2021image,xu2021predict}, image captioning \cite{hessel2021clipscore,mokady2021clipcap}, view synthesis \cite{jain2021putting}, object detection \cite{gu2021open,zhong2021regionclip,shi2022proposalclip}, and semantic segmentation \cite{rao2021denseclip,zhou2021denseclip}. 
These applications mainly focus on building the semantic relationship between texts and visual entities, and hence they suffer less from linguistic ambiguity. Different from existing works, we focus on the effectiveness of CLIP in understanding the quality of an image. The linguistic ambiguity in this task and the constant shape constraint of CLIP limit the performance of CLIP in perception assessment. In this work, we introduce two modifications to attenuate such limitations. Our resultant model CLIP-IQA outperforms the vanilla CLIP by a significant margin.